\documentclass{bmvc2k}

\title{Compositional Concept Erasure in Text-to-Image\\
       Diffusion Models via Hierarchically Grounded Semantic Surgery}

\addauthor{Chen Dai}{}{1}
\addauthor{Ganyu Zou}{}{1}
\addauthor{Nathan Self}{}{1}
\addauthor{Kevin Piper}{}{2}
\addauthor{Ramachandra Rao Seethiraju}{}{2}
\addauthor{Karthik Shyamsunder}{}{2}
\addauthor{Chang-Tien Lu}{}{1}
\addauthor{Naren Ramakrishnan}{}{1}

\addinstitution{
 Department of Computer Science\\
 Virginia Tech\\
 Alexandria, Virginia, USA
}
\addinstitution{
 Verisign, Inc.\\
 Reston, Virginia, USA
}

\runninghead{Dai et al.}{Compositional Concept Erasure via HGSS}

\usepackage{amsmath,amssymb}
\usepackage{algorithm}
\usepackage{algpseudocode}
\usepackage{graphicx}
\usepackage{booktabs}
\usepackage{multirow}
\usepackage{xcolor}
\usepackage{makecell}
\usepackage{array}
\usepackage{pifont}
\usepackage{url}
\usepackage{tikz}
\usetikzlibrary{arrows.meta,positioning,calc,shapes.geometric,fit}

\newcommand{\Enc}{\mathcal{E}}
\newcommand{\Aerase}{\mathcal{A}_E}
\newcommand{\Apreserve}{\mathcal{A}_P}
\newcommand{\KLsym}{\mathrm{KL}_{\mathrm{sym}}}
\newcommand{\HGSS}{HGSS}

\renewcommand{\SS}{SS}

\begin{document}
\maketitle

\begin{figure}[t]
\centering
\includegraphics[width=0.74\textwidth]{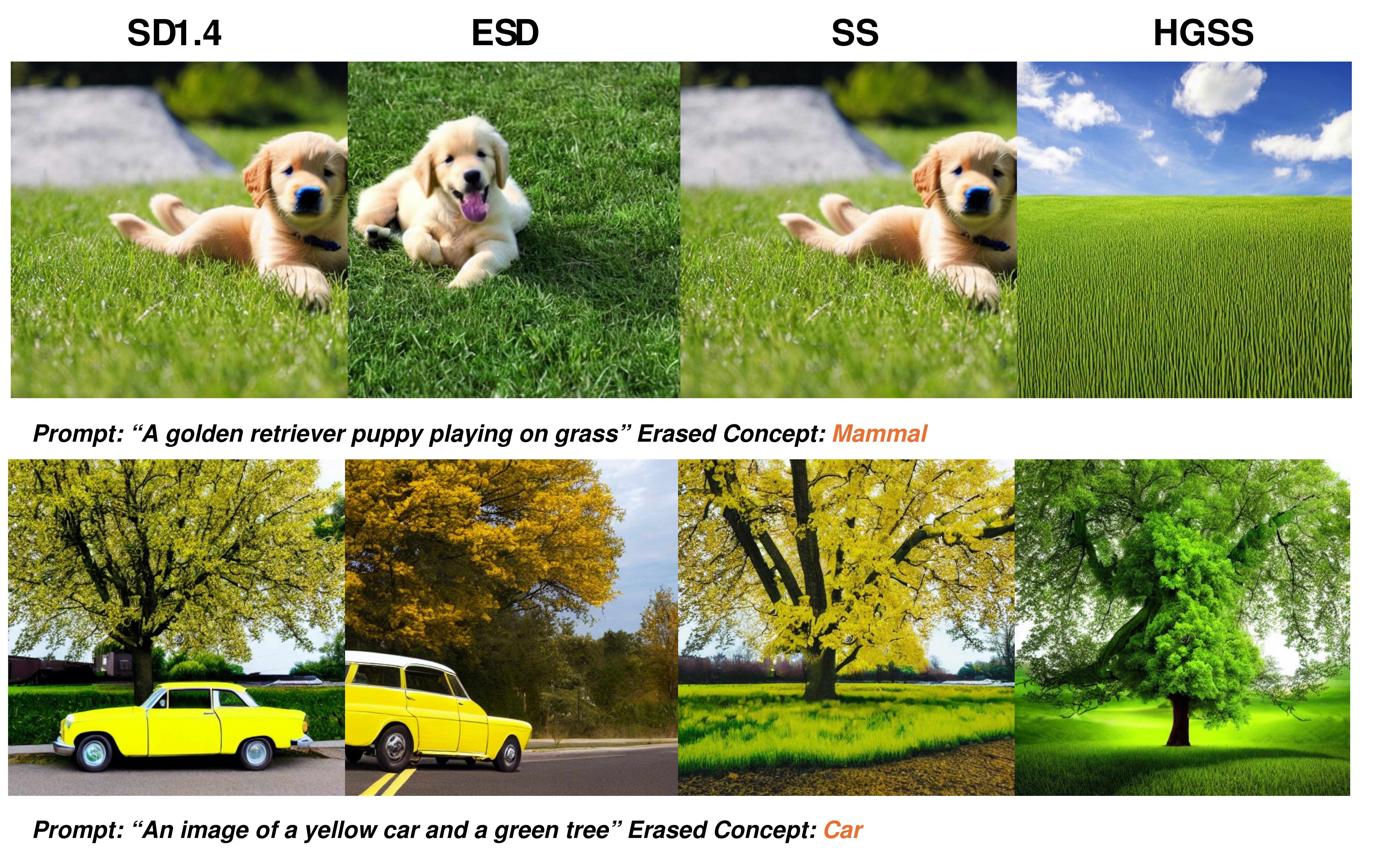}
\caption{\textbf{Two motivating failures in compositional concept
erasure.} Top: under erase target ``mammal'', prior methods can
retain a hierarchical descendant. Bottom: under erase target
``car'', ESD can retain the target, while \SS{} removes it but
transfers the target attribute to the preserved tree. \HGSS{}
removes the target concept while preserving non-target content.}
\label{fig:teaser}
\end{figure}

\begin{abstract}
Removing copyrighted, unsafe, or user-specified concepts from a
deployed text-to-image diffusion model is now a practical
requirement.
Weight-editing methods can suppress fixed targets, but they require
per-target retraining and modify the model checkpoint. Training-free
methods, on the other hand, are deployment-friendly, but they
suffer from text-side routing failures on compositional prompts.
In such prompts, the erase target may be invoked through a related
class rather than its lexical name, and its modifiers may migrate
onto preserved objects.
This paper proposes Hierarchically Grounded Semantic Surgery
(\HGSS{}), a training-free framework for compositional concept
erasure. The framework lifts both the routing signal and the edit
operator used by text-side erasure. First, hierarchical span
grounding resolves erase-target spans through lexical, taxonomic,
and semantic evidence, while guarding against broad-hypernym and
compound-head false positives. Second, dynamic attribute binding
refines the text conditioning during early denoising via a
counterfactual reference and a preserve-aware cross-attention
objective, keeping surviving attribute--noun bindings intact.
\HGSS{} selectively removes the erase target without updating
model weights or adding learned parameters.
On SEE, \HGSS{} cuts hierarchical evasion from $29.54$ to $10.02$
and roughly halves pairwise attribute leakage, achieving the best
Neighbor $E$ and Attr$P$ scores among the reported erasure
methods. On
UnlearnCanvas, \HGSS{} slightly improves the six-metric average
over the matched \SS{} baseline, reaching state-of-the-art.
\end{abstract}

\section{Introduction}
\label{sec:intro}

Removing copyrighted, unsafe, or user-specified concepts from a
deployed text-to-image diffusion model~\cite{rombach2022ldm} is now
a practical requirement. Real prompts, however, are compositional.
A target may be invoked through hierarchical or implicit reference,
and modifiers often remain bound to objects that the erasure
procedure is trying to remove. The requirement is especially sharp
for \emph{training-free} erasure, where the deployed U-Net is left
untouched and the intervention happens only through the
text-conditioning stream.

Existing methods occupy two deployment regimes. \emph{Weight-editing}
methods---ESD~\cite{gandikota2023esd}, UCE~\cite{gandikota2024uce},
MACE~\cite{lu2024mace}, and related approaches
(Sec.~\ref{sec:related})---can be strong on fixed targets. However,
they modify the model or attach learned modules, and each edit
incurs a training, storage, and distribution-shift cost.
\emph{Training-free} methods---SLD~\cite{schramowski2023sld},
SAFREE~\cite{yoon2025safree}, and Semantic Surgery
(\SS)~\cite{xiong2025ss}---are lighter: the same checkpoint can
serve many erase targets on demand. Their bottleneck is that the
edit must be computed from the prompt itself. Any routing error in
the text side therefore becomes an erasure failure.

\SS{}~\cite{xiong2025ss}, a strong training-free representative,
detects erase-related prompt terms via a similarity-based
concept-presence kernel and subtracts a weighted direction from the
conditioning hidden states. This handles the literal
safety--quality slice of I2P~\cite{schramowski2023sld}. Two failure
modes, however, become apparent on the compositional prompts of the
recent SEE benchmark~\cite{see2025}, and neither is reliably
addressed by tuning edit strength:

\begin{enumerate}\itemsep0pt
  \item \textbf{Hierarchical evasion.} When the prompt does not
        lexicalize the erase target, similarity-based selection
        misses descendants of the target. For example, the prompt
        ``a poodle running in the park'' under target ``mammal''
        still produces a mammal, because ``poodle'' is not
        lexically matched. The parent-to-child case is ambiguous
        in the opposite direction: ``flower'' may instantiate the
        fine-grained target ``rose'', and routing the edit through
        the parent risks suppressing non-target flowers. On the
        SEE evasion axis, \SS{} remained vulnerable, with a
        $29.54$ average despite strong neighboring erasure.
  \item \textbf{Attribute leakage.} Cross-attention text-side
        subtraction projects an aggregate direction and cannot
        determine whether a modifier such as ``blue'' belongs to
        the target noun or to a preserved one. As a result,
        attributes of the target can migrate onto preserved
        objects: ``a blue car next to a dog'' becomes a blue
        \emph{dog}. On the SEE attribute-leakage axis, \SS{}'s
        pairwise CLIP Attr$E$ exceeded twice that of weight-edit
        baselines (Attr$P$ was roughly $1.4\times$ theirs).
\end{enumerate}

These two failures, visualised in Figure~\ref{fig:teaser}, share a
structural cause. \SS{} treats the
prompt as a set of candidate terms and routes a single concept
direction through lexical--CLIP similarity. This primitive
represents neither the hierarchy that relates a target to its
descendants nor the binding that ties an attribute to its noun.
A larger \SS{} subtraction can sharpen the edit on lexicalized
targets. It cannot, however, recover an unselected descendant
token or decide which noun a modifier should remain bound to. The
remedy must therefore replace the structural primitives rather
than re-tune the existing scalars.

\paragraph{Contributions.}
This paper proposes Hierarchically Grounded Semantic Surgery
(\HGSS{}), a training-free framework for compositional
concept erasure. Its contributions are:
\begin{enumerate}\itemsep0pt
  \item \textbf{A structural framing for compositional erasure.}
        Existing training-free erasure methods treat prompt terms
        as loosely related routing cues. \HGSS{} instead views
        concept erasure as the inverse objective to compositional
        generation. The framework keeps the \SS{} coarse edit
        interface, but replaces its lexical routing and static
        conditioning primitives. The two replacements remain
        training-free: neither introduces learned parameters nor
        per-concept hand-coded rules.
  \item \textbf{Hierarchical span grounding.} To decide what to
        edit, \HGSS{} introduces a resolver for erase-target spans
        (Sec.~\ref{sec:resolver}). The resolver combines lexical
        matches, WordNet-backed taxonomic evidence, and semantic
        fallback. Broad-hypernym and compound-head guards reduce
        false positives on preserved objects.
  \item \textbf{Dynamic attribute binding.} To decide how the edit
        evolves, \HGSS{} introduces a per-step conditioning update
        for early denoising (Sec.~\ref{sec:dyntext}). A
        counterfactual reference defines the target-removed state.
        An asymmetric attention objective protects preserve spans
        and suppresses leakage from erase-target spans.
  \item \textbf{Evaluation across complementary axes.} The
        experiments evaluate the two structural mechanisms on
        compositional side effects, style/object unlearning, and
        direct lexical safety compatibility. Across the three
        axes, the resolver and dynamic binding reduce compositional
        side effects while broadly preserving image quality and
        the safety capability of the inherited coarse edit.
        Ablations separate the resolver from dynamic binding
        (Sec.~\ref{sec:exp}).
\end{enumerate}
Taken together, the two structural mechanisms view erasure as an
\emph{inverse objective to compositional generation}. Rather than
binding every prompt entity correctly, the framework first
routes the erase target and its descendants into the edit set,
then stabilises the surviving preserve-side bindings during
denoising. The grounding cascade and the dynamic-binding update
are implemented as a plug-in module for training-free, text-side
concept erasure.

\section{Related Work}
\label{sec:related}

\paragraph{Parameter-modifying concept erasure.}
A large body of work edits the diffusion model itself to remove a
target concept. \emph{Fine-tuning objectives} steer the generator
away from the target distribution.
ESD~\cite{gandikota2023esd} fine-tunes U-Net attention with a
negative-guidance loss, Concept Ablation~\cite{kumari2023ca}
matches a surrogate concept, and
AdvUnlearn~\cite{zhang2024advunlearn} optimises adversarial
robustness during erasure. \emph{Closed-form and low-rank weight
edits} act directly on cross-attention projections without full
fine-tuning. UCE~\cite{gandikota2024uce} and
RECE~\cite{gong2024rece} apply closed-form edits,
MACE~\cite{lu2024mace} targets many concepts jointly, and
SPM~\cite{lyu2024spm} uses one-dimensional adapters.
\emph{Sparsity- and saliency-guided edits} localise the
intervention to a subset of parameters. SalUn~\cite{fan2024salun}
uses gradient-based weight saliency,
SAeUron~\cite{cywinski2025saeuron} edits sparse-autoencoder
features, and ConceptPrune~\cite{chavhan2025conceptprune} prunes
skilled neurons. \emph{Locality-oriented and representation-space
methods} reduce collateral damage by constraining where the edit
applies. CPE~\cite{lee2025cpe} introduces residual attention
gates, GLoCE~\cite{lee2025gloce} gates a training-free low-rank
adaptation, and AdaVD~\cite{wang2025valuespace} edits the value
space along orthogonal complements. A further line refines
\emph{target geometry}: AGE~\cite{bui2025age} identifies which
concepts to push erased samples toward, while
CoGFD~\cite{nie2025cogfd} explicitly handles combinations of
erase targets rather than single nouns. These methods exploit
additional parameter capacity to retain preserve-side fidelity,
but require per-target training and modified checkpoints.

\paragraph{Training-free and inference-time erasure.}
Training-free erasure preserves the base generator and intervenes
only at sampling time. SLD~\cite{schramowski2023sld} modifies
classifier-free guidance using a safety prompt;
SAFREE~\cite{yoon2025safree} projects text embeddings away from
unsafe directions before they enter the U-Net.
TraSCE~\cite{jain2024trasce} steers the denoising trajectory away
from harmful regions via a localised loss-based guidance.
Semantic Surgery (\SS)~\cite{xiong2025ss} subtracts a
similarity-weighted concept direction from the conditioning hidden
states. These methods are deployment-friendly since the same
checkpoint serves many erase targets on demand, but their edits
are only as reliable as the text-side signals that select the
edited span. When the target is reached through a related class,
a semantic variant, or an entangled attribute phrase, a single
conditioning-time subtraction can fail in two ways. It can miss
the routing path entirely, or remove content beyond the intended
span.

\paragraph{Side effects and robustness.}
Early erasure evaluations emphasise literal unsafe-content
removal and clean-set image quality, exemplified by
I2P~\cite{schramowski2023sld} with NudeNet-based scoring.
UnlearnCanvas~\cite{zhang2024unlearncanvas} expands this view to
style and object unlearning across $60$ styles and $20$ object
classes.
SEE~\cite{see2025} factors side effects into \emph{neighboring
impact}, \emph{hierarchical evasion}, and \emph{attribute
leakage}, each with its own protocol. A complementary line
stress-tests erased models with adversarial
prompts: Ring-A-Bell~\cite{tsai2024ringabell},
UnlearnDiffAtk~\cite{zhang2024unlearndiffatk},
MMA-Diffusion~\cite{yang2024mmadiffusion}, and
P4D~\cite{chin2024p4d}.

\paragraph{Text-side compositional control.}
A separate line edits text conditioning states during sampling to
repair compositional generation failures. Prompt-to-Prompt
\cite{hertz2023p2p} swaps cross-attention maps between related
prompts.
Attend-and-Excite~\cite{chefer2023attendexcite} strengthens the
attention of subject tokens that risk being neglected.
Structure Diffusion~\cite{feng2023structurediffusion} incorporates
linguistic structure into the guidance computation, and
DATE~\cite{na2025date} updates the text embedding per denoising
step to correct attribute--object misbinding.
SynGen~\cite{rassin2023syngen} introduces a positive--negative
cross-attention loss that aligns each attribute's attention with
the noun it modifies.

\HGSS{} sits at the intersection of these lines by keeping the
deployment advantage of training-free erasure but replacing the
similarity-only routing used by prior inference-time methods with
a hierarchical span grounding cascade. From text-side
compositional control, \HGSS{} reuses attribute--object binding
primitives on the preserve side. \HGSS{} inverts that framework's
objective by first routing target spans and descendants into
the edit set, then re-applying binding control only to surviving
spans, thereby treating concept erasure
as the \emph{inverse objective to compositional generation},
motivating the counterfactual reference and preserve-restricted
attention pair in Sec.~\ref{sec:dyntext}.

\section{Method}
\label{sec:method}

\subsection{Preliminaries and Notation}
\label{sec:prelim}

\HGSS{} is described here in the setting of a latent text-to-image
diffusion model with a frozen autoencoder, a CLIP text
encoder~\cite{radford2021clip}~$\Enc$, and a U-Net
$\epsilon_\theta$ trained for
$\epsilon$-prediction~\cite{ho2020ddpm}. Given a prompt $p$, the
text encoder returns conditioning hidden states
$c = \Enc(p) \in \mathbb{R}^{L\times d}$. At inference, a
DDIM~\cite{song2021ddim} schedule iteratively denoises a latent
$z_T$ over $T{=}50$ steps using classifier-free guidance
(CFG)~\cite{ho2022cfg} with scale $w$. In our experiments, this
instantiation is Stable Diffusion v1.4~\cite{rombach2022ldm}.

A concept-erasure method takes $c$, an erase target $T_e$, and an
optional preserve target $T_p$. It returns an edited conditioning
$\tilde c$ such that the sampled image no longer depicts $T_e$
while the non-target content remains intact. The
\SS{}~\cite{xiong2025ss} formulation serves as the base: it scores
candidate prompt terms against the prompt context through a
sigmoid concept-presence kernel parameterised by an inherited
offset $\beta$ and temperature $\gamma$ (see
App.~B.1 for the kernel and threshold
values used). The selected weighted spans
$\{(t_k,\omega_k)\}_{k\in S}\subseteq\Aerase$
(Sec.~\ref{sec:resolver}) then drive
a single weighted concept-direction subtraction,
\begin{equation}
\label{eq:ss-edit}
\tilde c \;=\; c \;-\; \alpha_{\star}
   \sum_{k\in S}\hat\omega_k\,
   \big(\Enc(t_k) - \Enc(\varnothing)\big),
\end{equation}
with $\hat\omega_k = \omega_k/\sum_{j\in S}\omega_j$, $\varnothing$
the empty prompt, and aggregate edit coefficient $\alpha_\star$.
In the \SS{} base, selection is similarity-driven and spans
contribute uniformly. When preserve spans are grounded, the
aggregated erase direction is projected onto the orthogonal
complement of the weighted preserve direction before subtraction,
at a fixed unit scale. \HGSS{}
determines \emph{which spans enter $S$ and their weights} and
\emph{how the edited conditioning evolves during sampling}, and
does not introduce any new scalar at the coarse-edit interface.

\subsection{Overview of \HGSS}
\label{sec:overview}

\begin{figure}[t]
\centering
\includegraphics[width=0.82\textwidth]{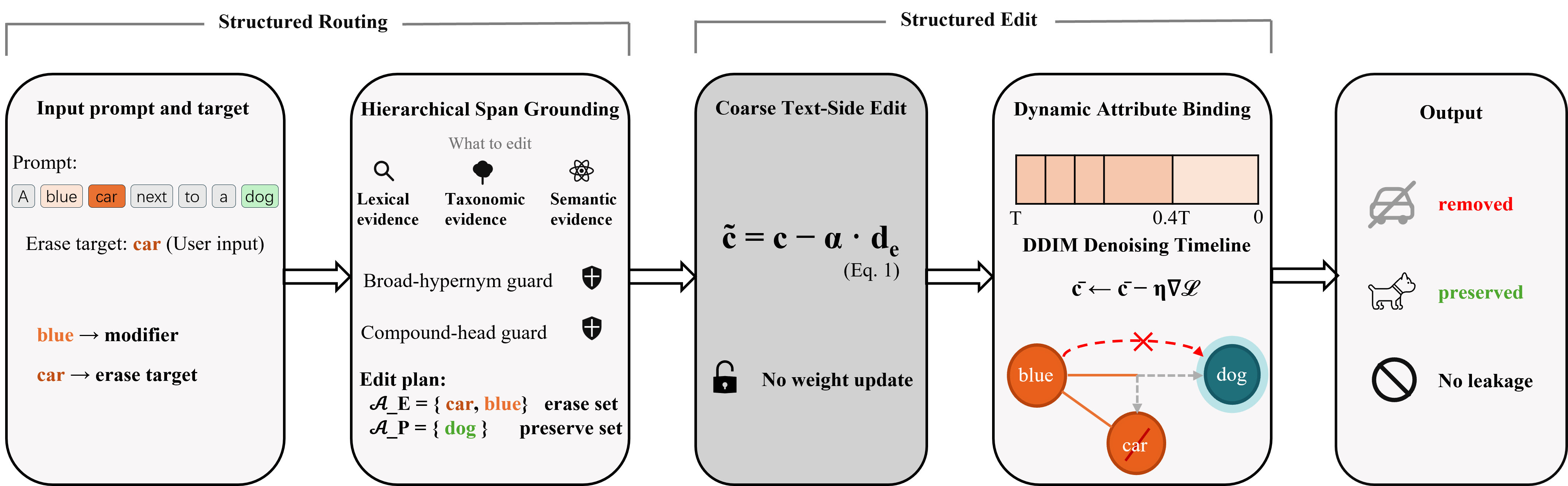}

\caption{\textbf{\HGSS{} pipeline.}
\HGSS{} surrounds the \SS{}-inherited weighted coarse edit
(Eq.~\eqref{eq:ss-edit}) with two training-free modules.
\emph{Hierarchical Span Grounding} produces the weighted
edit plan from lexical, taxonomic, and semantic evidence, gated by
broad-hypernym and compound-head guards
(Sec.~\ref{sec:resolver}). \emph{Dynamic Attribute Binding}
stabilises the preserve-side bindings exposed by the
resolver-expanded erase set, refining the live conditioning during
the first $0.6$ fraction of DDIM iterations via an asymmetric
preserve-aware cross-attention objective (Sec.~\ref{sec:dyntext}). The U-Net
and the CLIP text encoder remain frozen throughout.}
\label{fig:pipeline}
\end{figure}

\HGSS{} makes two structural changes to the coarse-edit pipeline,
which form two halves of the framework.
\textbf{Structured routing} decides
\emph{what to edit} by using hierarchical span grounding instead of 
token-level similarity (Sec.~\ref{sec:resolver}).
\textbf{Structured edit} decides \emph{how the edit
evolves} via dynamic attribute binding during
denoising, rather than a single conditioning subtraction
(Sec.~\ref{sec:dyntext}). The two modules address hierarchical
evasion and
attribute leakage asymmetrically:
structured routing supplies the main erase-side correction, while
structured edit stabilises preserve-side bindings so the routing
gain is not paid for by leakage onto surviving nouns. The two
halves and the \SS{}-inherited coarse edit between them comprise the
visual pipeline in Figure~\ref{fig:pipeline}.

Given $(p, T_e, T_p)$, \HGSS{} executes three inference stages
(Figure~\ref{fig:pipeline}; full algorithm in
App.~A.3):
\begin{enumerate}\itemsep0pt
\item \textbf{Ground.} The span-grounding module returns weighted
      erase spans $\Aerase$ and preserve spans $\Apreserve$ in the
      prompt, using lexical, hierarchical, and semantic evidence.
\item \textbf{Coarse edit.} The coarse edit
      (Eq.~\eqref{eq:ss-edit}) is applied using the grounded erase
      spans when available and falls back to the base
      similarity-driven selection otherwise, producing the static
      anchor $\tilde c$.
\item \textbf{Bind.} A text-side localization module builds
      counterfactual references and preserve-side attribute
      anchors; during denoising, the dynamic update refines the
      live conditioning $\bar c$ with attention and CLIP losses.
\end{enumerate}

\subsection{Hierarchical Span Grounding}
\label{sec:resolver}

This module addresses \emph{hierarchical evasion}: given
prompt $p$ and target $T_e$, it returns the prompt spans that
should be treated as erase-positive evidence, including spans that
are hierarchical descendants of $T_e$ when $T_e$ itself does not
appear lexically in
$p$. \SS{} solves this with CLIP similarity alone, which is
unreliable when $T_e$ is not lexicalized. \HGSS{} replaces this
similarity-only test with a deterministic cascade that emits
weighted \emph{(span, relation, confidence)} tuples and removes false positives via guard rules so that the remaining tuples become the
weighted erase set $\Aerase$ and, by complement, the preserve set
$\Apreserve$.

\paragraph{Cascade.}
The cascade consults \emph{three categories of evidence} in order of signal reliability: lexical,
taxonomic, and semantic. Lexical
evidence provides high precision. Taxonomic evidence expands recall
when lexical matches fail. And, semantic fallback handles
open-vocabulary concepts beyond the lexicon. These three categories
are visualised as the three labelled icons in Box~2 of
Figure~\ref{fig:pipeline} and are realised through four sources,
consulted in priority order. The first source that emits a
confident candidate determines the relation label.
\textbf{(1) Lexical evidence} tests exact equality, substring
containment, and tokenized head-noun match between prompt term
$t_k$ and $T_e$. Exact matches are labelled \textsc{self}.
\textbf{(2a) Taxonomic backoff} uses
WordNet~\cite{miller1995wordnet} to supply non-lexical evidence
when lexical matching fails. The resolver looks up synsets for
$t_k$ and $T_e$ and runs a breadth-first traversal of the
hypernym/hyponym
directed acyclic graph (DAG) with max depth $12$. The closest valid relation is classified as
\textsc{descendant}/\textsc{child} (force-include),
\textsc{parent}/\textsc{parent-like}, or \textsc{sibling} via shared
hypernyms. A \emph{broad-hypernym guard} suppresses generic
ancestors (\texttt{entity}, \texttt{object}, \texttt{whole},
\texttt{artifact}, \texttt{organism}, \dots) so that ``car'' and
``apple'' do not become hierarchically related through the root of
the WordNet DAG.
\textbf{(2b) Activity bridge} (a taxonomic refinement) handles
verb-style targets and gerund-derived identifiers by expanding each
candidate through WordNet derivational links into verb, gerund, and
nominal variants, then applying the same relation and CLIP checks.
The expansion is generated per prompt rather than taken from a
hand-curated activity lexicon.
\textbf{(3) Semantic fallback} admits a candidate when its
CLIP cosine similarity to $T_e$ exceeds a stage-specific threshold
(distinct values for primary erase, derived modifiers, and activity
targets; full settings are in App.~B.1).
This stage handles out-of-WordNet concepts (style names, brand
names) without hand-curated lists.

\paragraph{Guards.}
Two guard rules reduce the most common false-positive patterns in
our development set:
\begin{itemize}\itemsep0pt
\item \textbf{Compound-head guard.} For a compound noun
      $t_k = w_1\,w_2\,\ldots\,w_n$, the derived head $w_n$ is only
      erased if the full compound $t_k$ has an erase-positive
      relation to $T_e$. A relation between only $w_n$ and $T_e$ in
      isolation will not necessarily trigger erasure.
      This prevents ``\underline{baseball bat}'' from triggering
      erasure under a generic ``mammal'' target: the derived head
      ``bat'' is a mammal in WordNet, but the full compound
      ``baseball bat'' does not bear a \texttt{descendant}
      relation to \emph{mammal}.
\item \textbf{Self-suppression.} If the prompt lexicalizes a
      descendant of $T_e$ (the resolver returns a
      \textsc{descendant}/\textsc{child} candidate), the \textsc{self}
      candidate corresponding to $T_e$ itself is suppressed to avoid
      \emph{double-counting} when $T_e$ is also a prompt token.
\end{itemize}

\paragraph{The edit plan.}
The resolver returns a weighted erase set
$\Aerase=\{(t_k,\omega_k)\}_k$ and an analogously weighted
preserve set $\Apreserve$.
The weight $\omega_k$ reflects the cascade's confidence in
candidate $t_k$: force-included relations (\textsc{self},
\textsc{descendant}, \textsc{child}) contribute at a default
full weight; threshold-admitted candidates contribute at their
cascade-stage confidence score. All other candidates must clear
a similarity-confidence threshold and a separate preserve-strength
floor that protects against weakly related nouns. When the user
provides an explicit preserve target $T_p$ (as in SEE and
UnlearnCanvas protocols), the resolver anchors $T_p$'s span in
$\Apreserve$; otherwise $\Apreserve$ is the complement of
$\Aerase$ within the prompt. All threshold values were fixed
once during development and not tuned on SEE or UnlearnCanvas;
exact values are listed in App.~B.1.
The role separation between span grounding and dynamic binding is
evaluated through module-level diagnostic ablations in
Sec.~\ref{sec:exp-ablate}.

\paragraph{Worked example.}
Under target ``mammal'', the resolver marks ``poodle'' (in
``a poodle running in the park'') as a force-included
descendant span via WordNet backoff. The compound-head
guard, however, suppresses ``bat'' in ``a baseball bat on the
kitchen counter'' because the full compound does not bear an
erase-positive relation. The full cascade trace is in
App.~A.1.

\subsection{Dynamic Attribute Binding}
\label{sec:dyntext}

Dynamic Attribute Binding is not intended to be a standalone
erasure mechanism. It refines the live conditioning during early
DDIM steps so that, after the resolver-grounded coarse edit, the
modifier--noun bindings on spans that should survive the erasure
remain intact. A small gradient correction is applied to the
conditioning embeddings during early DDIM steps, when object
layout and attribute binding are still being established. The
update combines two parts: a per-token counterfactual reference
and a per-step composite loss. The reference anchors what each
token position should encode after $T_e$ is removed. The
composite loss pulls each preserved (modifier, head-noun) pair
together while pushing preserve-side keys away from the residue
left by the coarse edit.

\paragraph{Design rationale.}
The four loss components in Eq.~\eqref{eq:date-loss} (introduced
below) are not interchangeable. Each constrains a distinct level
of the conditioning pipeline. The update concentrates in early
denoising, where object layout and cross-attention bindings are
most malleable. Detailed motivation for each component (the
symmetric-KL choices for the attention losses, the residual
CLIP-space target penalty, and the $\ell_2$ prior anchor) is in
App.~A.2.

\subsubsection{Counterfactual Reference Encoding}
\label{sec:cf}
For each erase span, the prompt is re-encoded with that span
replaced by the CLIP PAD token, producing a counterfactual hidden
$\Enc(p^{-T_e})$. This induces a per-token counterfactual reference
\begin{equation}
\label{eq:cf-direction}
\bar c^{\mathrm{ref}}_k \;=\;
\begin{cases}
\Enc(p^{-T_e})_k, & \text{$k$ in an erase span},\\[2pt]
\Enc(p)_k,        & \text{otherwise},
\end{cases}
\end{equation}
which acts as a fixed anchor for two \emph{deterministic
hidden-state operations} applied before the gradient update of
Sec.~\ref{sec:date-update}. (i) At the erase positions, the live
embedding is substituted with $\bar c^{\mathrm{ref}}$ at a
configurable strength, so the conditioning moves toward what the
prompt would encode without $T_e$. (ii) At the preserve positions,
a soft pull toward $\bar c^{\mathrm{ref}}$ is applied so the
surrounding modifiers and head nouns are protected against drift
caused by the erase substitution.
A token-norm gate ($\|\Enc(p)_k-\Enc(p^{-T_e})_k\|_2 < 10^{-5}$
treated as zero) skips spans where the encoder produces no
counterfactual displacement.

\subsubsection{Per-Step Text-Conditioning Refinement}
\label{sec:date-update}

Let $i \in \{0, 1, \dots, T{-}1\}$ index the DDIM sampling
iterations, and let $t_i$ denote the scheduler timestep at
iteration $i$ (so $i{=}0$ is the highest-noise step and $i{=}T{-}1$
the lowest). Every $\Delta t {=} 10$ iterations during the
\emph{early-denoising window} (the first $t_{\text{end}}{=}0.6$
fraction of the sampling trajectory, i.e.\ $i/(T{-}1) \le
t_{\text{end}}$), the live conditioning embedding $\bar c$ is
updated via a single gradient descent step on a composite
attention and CLIP loss
\begin{equation}
\label{eq:date-loss}
\mathcal{L}^{(t)} =
   \lambda_{\text{bind}}\,\mathcal{L}_{\text{bind}}^{(t)}
 + \lambda_{\text{neg}} \,\mathcal{L}_{\text{neg}}^{(t)}
 + \lambda_{\text{tgt}} \,\mathcal{L}_{\text{tgt}}^{(t)}
 + \lambda_{\text{prior}} \,\mathcal{L}_{\text{prior}}^{(t)},
\end{equation}
with step size $\eta{=}0.5$ and weights
$(\lambda_{\text{bind}}, \lambda_{\text{neg}}, \lambda_{\text{tgt}},
\lambda_{\text{prior}}) {=} (0.4,\,0.1,\,1.0,\,0.02)$ fixed across
all experiments. The four components are:

\paragraph{(a) Preserve-binding attention loss
$\mathcal{L}_{\text{bind}}^{(t)}$.}
The first term keeps each preserved attribute attentionally tied
to its noun, so that the coarse subtraction in
Eq.~\eqref{eq:ss-edit} does not detach modifiers from the nouns
that survive. Let $\mathbf{A}^{(t)}_k$ be the cross-attention
probability map over query positions for prompt token $k$ at step
$t$, and let $\mathcal{B}$ denote the set of (modifier,
head-noun) token-position pairs extracted from the preserve
spans by the resolver (Sec.~\ref{sec:resolver}). Each
pair's attention is pulled together via symmetric KL,
\begin{equation}
\mathcal{L}_{\text{bind}}^{(t)} \;=\;
\!\!\!\sum_{(a,n)\in\mathcal{B}}\!\!\!
   \rho_{a,n}\,
   \KLsym\!\big(\mathbf{A}^{(t)}_a,\,
                 \mathbf{A}^{(t)}_n\big),
\label{eq:bind}
\end{equation}
where $\rho_{a,n}$ is the inverse modifier--noun token distance.
This is the cross-attention analogue of compositional
attribute--object binding~\cite{rassin2023syngen}, applied
\emph{selectively} to the spans the resolver marks as preserved.

\paragraph{(b) Leakage-suppression loss
$\mathcal{L}_{\text{neg}}^{(t)}$.}
Building on the positive/negative attention principle of
SynGen~\cite{rassin2023syngen}, the second term pushes each
preserve attribute key away from the rest of the prompt. This
prevents its attention mass from drifting onto neighbouring
positions left over by the coarse edit. For each preserve anchor
$(a,n)\!\in\!\mathcal{B}$, let $\mathcal{S}_{a,n}{=}\{a,n\}$ be the
source keys and $\mathcal{K}_{\text{out}}$ the set of prompt tokens
outside the anchor subtree. The symmetric KL between source-key
attention and the non-subject distribution is \emph{maximised},
\begin{equation}
\mathcal{L}_{\text{neg}}^{(t)} \;=\;
-\!\!\!\!\sum_{(a,n)\in\mathcal{B}}\,
\sum_{s \in \mathcal{S}_{a,n}}\!\!
\KLsym\!\big(\mathbf{A}^{(t)}_s,\,
              \mathbf{A}^{(t)}_{\mathcal{K}_{\text{out}}}\big).
\label{eq:neg}
\end{equation}
Combined with (a), the two attention losses constitute an
\emph{asymmetric} attention objective by aligning the bindings that the prompt
specifies, and separating them from the keys that the erase-target
span leaves behind. Together with the resolver-expanded
$\Aerase$, this is the principal \emph{preserve-side} mechanism
by which \HGSS{} prevents the coarse subtraction from detaching
modifiers from surviving nouns.

\paragraph{(c)--(d) Target and prior terms.}
$\mathcal{L}_{\text{tgt}}^{(t)}$ is a CLIP-margin penalty that
lowers the predicted image's similarity to $T_e$ relative to
neutral captions, providing a residual sample-level safety signal.
$\mathcal{L}_{\text{prior}}^{(t)}
= \|\bar c_{\mathrm{live}} - \tilde c_{\mathrm{anchor}}\|_2^2$
anchors the live conditioning $\bar c_{\mathrm{live}}$ to the
frozen coarse-edit anchor $\tilde c_{\mathrm{anchor}}$
(Eq.~\eqref{eq:ss-edit}), preventing long-horizon drift. The
update target is $\bar c_{\mathrm{live}}$, with per-sample gradient
clipping. Full
formulations, the latent-space update variant, and design
rationale are in App.~A.2.

\subsection{Inference and Complexity}
\label{sec:algo}

At inference, the resolver (Sec.~\ref{sec:resolver}) returns
$\Aerase$ and $\Apreserve$; the coarse edit (Eq.~\eqref{eq:ss-edit})
produces the static anchor $\tilde c$; during the first
$t_{\text{end}}{=}0.6$ fraction of DDIM steps, the live
conditioning $\bar c$ is refined by Eq.~\eqref{eq:date-loss} every
$\Delta t{=}10$ iterations; the U-Net is otherwise sampled by
standard DDIM. With $T{=}50$, the refinement fires at
$i\in\{0,10,20\}$, constituting three backward passes through the frozen
model per image, with no additional checkpoint storage beyond the
resolver's WordNet cache. The full algorithm is in
App.~A.3.

\section{Experiments}
\label{sec:exp}

\subsection{Experimental Setup}
\label{sec:exp-setup}

\paragraph{Base model and inference.}
\HGSS{} and the training-free baselines are evaluated as drop-in
text-side modifiers of Stable
Diffusion v1.4~\cite{rombach2022ldm} with the DDIM sampler at $50$
inference steps, classifier-free guidance scale $7.5$, image
resolution $512{\times}512$. No fine-tuning or distillation is
performed. For controlled comparison, all \SS{} baselines are
evaluated with visual feedback \emph{disabled} (no-VF). \HGSS{}
does not use visual feedback, so the \SS{}$\rightarrow$\HGSS{}
comparison isolates the two proposed modules under the same
coarse text-side edit interface. All reported \SS{} and \HGSS{}
rows use the same evaluation protocol and the same
benchmark-specific coarse-edit operating point. For compositional
erasure (SEE, UnlearnCanvas) the \SS{} coarse-edit operating
point is $\beta{=}{-}0.12$; for direct safety erasure (I2P) it is
$\beta{=}{-}0.06$, matching the original \SS{} safety operating
point. In every setting \emph{\HGSS{} introduces no new
hyperparameter at this interface and runs at exactly the same
$\beta$ as the matched \SS{} baseline}.
Per-step refinements use the window $[0,\,0.6T]$ and interval
$\Delta t {=} 10$. The step size is $\eta{=}0.5$. The loss
weights are
$(\lambda_{\text{bind}}, \lambda_{\text{neg}}, \lambda_{\text{tgt}},
\lambda_{\text{prior}}) {=} (0.4,\,0.1,\,1.0,\,0.02)$ of
Sec.~\ref{sec:date-update}; identical settings are used on every
benchmark, and the update target is the live conditioning embedding
$\bar c$. Under matched inference conditions, the measured
per-image wall-clock cost is approximately $1.3\times$ \SS{}'s on
a single L40S GPU at $512{\times}512$ resolution. This is
consistent with the
$\approx 3$ extra backward passes derived in Sec.~\ref{sec:algo}.

\paragraph{Benchmarks.}
Three complementary benchmarks were used.
\textbf{(i) SEE}~\cite{see2025} served as the primary diagnostic
benchmark. Its three side-effect axes isolate the compositional
failures targeted by \HGSS{}: \emph{neighboring impact} (SEE
Table~2), \emph{evasion} (SEE Table~3), and \emph{attribute
leakage} (SEE Table~4).
The benchmark spans $11$ super-categories with $79$ base objects
and $5{,}056$ compositional variants. The official CLIP-based
scoring protocol was followed and metrics were reported in the
paper direction ($\downarrow$ for erase/evasion/leakage;
$\uparrow$ for preserve).
Neighbor $E$ measures residual target evidence on erase-side
prompts, while Neighbor $P$ measures preservation of related
non-target concepts. Evasion measures target reappearance through
hierarchical or related-class prompts. Attr$E$ and Attr$P$ measure
pairwise attribute leakage on erase and preserve sides.
\textbf{(ii) UnlearnCanvas}~\cite{zhang2024unlearncanvas}
measures style and object unlearning quality. The evaluation
covers the $50$ style targets and $20$ object targets defined by
the official benchmark protocol (selected from the $60$-style
UnlearnCanvas dataset). The reported metrics are the standard
unlearning accuracy (UA), in-domain retention accuracy (IRA),
and cross-domain retention accuracy (CRA) triple, with the
six-metric average across style and object axes. FID was computed
using a regenerated reference set with five images per
combination. The \SS{} and \HGSS{} rows are therefore strictly
comparable.
UA measures target removal. IRA and CRA measure retention on
in-domain and cross-domain non-target generations.
\textbf{(iii) I2P}~\cite{schramowski2023sld} measures direct
safety erasure. Per-class NudeNet detection counts are taken
under the standard \texttt{paper8 + threshold=0.6} protocol on
all $4{,}703$ I2P prompts across eight body-part classes.
MS-COCO-30K FID and CLIP are reported as image-quality sanity
checks.

\paragraph{Compared methods.}
\HGSS{} was compared against $17$ prior methods spanning
weight-editing/fine-tuning and training-free families. The
per-benchmark method lists appear in
Tables~\ref{tab:see_main}, \ref{tab:uc_main}, and
\ref{tab:i2p_main} (full I2P list in Suppl.\ Table~2).
Prior-method numbers were taken from their
published benchmark tables, including the I2P collation in
\cite{xiong2025ss}; the \SS{} and \HGSS{} rows use matched runs
under the same evaluation protocol.

\subsection{Main Results on SEE}
\label{sec:exp-see-main}

The three SEE axes test whether \HGSS{} shifts the intended
structural failure modes, and not edit strength only.
Table~\ref{tab:see_main} aggregates these axes: neighboring
preserve/erase (SEE Table~2), average evasion across $11$
super-categories (SEE Table~3), and pairwise CLIP attribute-leakage
(SEE Table~4).

\begin{table}[!ht]
\centering
\footnotesize
\renewcommand{\arraystretch}{0.95}
\caption{\textbf{SEE aggregate results} in the original paper
direction. Neighbor $E$/$P$ metrics from SEE Table~2; Evasion
Avg.\ from SEE Table~3; Attr.\ Leakage columns from SEE Table~4
(pairwise CLIP without substring matching). Lower is better for
$E$/Evas./Leak., higher is better for $P$. Best \textbf{bold},
second-best \underline{underlined}, excluding the unedited
SD~v1.4 reference row. \SS{} and \HGSS{} rows use
matched runs under the same evaluation protocol.}
\label{tab:see_main}
\setlength{\tabcolsep}{4pt}
\begin{tabular}{lccccc}
\toprule
Method & Neighbor & Neighbor & Evas. & Attr$E$ & Attr$P$\\
       & $E$\,$\downarrow$ & $P$\,$\uparrow$ & Avg\,$\downarrow$
       & /CLIP\,$\downarrow$ & /CLIP\,$\downarrow$ \\
\midrule
Unedited SD v1.4~\cite{rombach2022ldm}      & 92.70 & 92.17 & 91.65 & 92.21 & 35.01 \\
\midrule
\multicolumn{6}{l}{\emph{Weight-edit / fine-tuning baselines}}\\
UCE~\cite{gandikota2024uce}                 & 30.00 & 66.85 & 84.25 & 31.56 & \underline{52.14} \\
RECE~\cite{gong2024rece}                    & \underline{23.08} & 57.62 & 83.49 & \textbf{24.43} & 57.26 \\
MACE~\cite{lu2024mace}                      & 28.68 & 55.47 & 81.34 & 29.33 & 58.87 \\
SPM~\cite{lyu2024spm}                       & 34.44 & 53.30 & 83.29 & 33.04 & 61.09 \\
ESD~\cite{gandikota2023esd}                 & 32.80 & 53.90 & 84.26 & 32.90 & 60.90 \\
AdvUnlearn~\cite{zhang2024advunlearn}       & 24.70 & 54.90 & 83.45 & \underline{27.50} & 60.00 \\
\midrule
\multicolumn{6}{l}{\emph{Training-free methods}}\\
\SS~\cite{xiong2025ss}                      & 25.30 & \textbf{31.22}\textsuperscript{\textdagger} & \underline{29.54} & 80.41 & 82.22 \\
\textbf{\HGSS{}} (ours)                     & \textbf{8.08}  & \underline{30.04}\textsuperscript{\textdagger}
                                                              & \textbf{10.02} & 41.44 & \textbf{38.30} \\
\bottomrule
\end{tabular}
\\[0.3em]
\parbox{0.96\linewidth}{\footnotesize
\textsuperscript{\textdagger}$P$/CLIP for training-free methods is
not directly comparable to the weight-edit rows (different
preserve-set construction in SEE Table~2); bold/underline in $P$
reflect best/second-best among training-free methods.}
\end{table}

\HGSS{} has the strongest score among erasure methods on
three columns:
Neighbor $E$/CLIP ($8.08$), evasion average ($10.02$), and Attr$P$
($38.30$). On Neighbor $E$, \HGSS{} more than halves the previous
overall best ($23.08$ from RECE~\cite{gong2024rece}). On evasion,
it cuts the matched \SS{} baseline from $29.54$ to $10.02$
($-66.1\%$ relative). On Attr$P$, it drops from $82.22$ to
$38.30$ ($-53.4\%$ relative) which is the largest
\SS{}$\rightarrow$\HGSS{} leakage reduction in
Table~\ref{tab:see_main}. On
Attr$E$, \HGSS{} reduces the matched \SS{} baseline from $80.41$
to $41.44$, but it does not close the gap to the strongest
weight-edit baselines: RECE ($24.43$) and AdvUnlearn ($27.50$).
This remaining gap is consistent with the training-free setting.
\HGSS{} leaves the U-Net frozen and intervenes only through
text-side conditioning. The relevant gain is therefore the large
matched improvement over \SS{}, not necessarily because of
weight editing on erase-side residual evidence. Preserve Neighbor
$P$/CLIP ($30.04$) stays within noise of the matched \SS{}
baseline ($31.22$). The Neighbor~$E$, Evasion, Attr$E$, and
Attr$P$ gains are all obtained at the same
$\beta$ as the matched \SS{} baseline, so the improvement
reflects the new mechanisms rather than more aggressive editing.

\paragraph{Per-category evasion breakdown.}
Table~\ref{tab:see_per_cat} reports the full SEE Table~3 (evasion)
breakdown. \HGSS{} achieves the strongest evasion score on every
one of the $11$ super-categories. Against weight-edit baselines
the gain is large in every category (at least $2\times$ lower
residual evasion everywhere, over $10\times$ in six of eleven
categories). Against the matched \SS{} baseline the
gain is consistent but smaller and varies by category. The largest
\SS{}$\rightarrow$\HGSS{} drops occur on Electronic
($35.16\to2.34$), Animal ($33.75\to1.56$), Sports
($30.47\to4.37$), and Furniture ($29.95\to4.69$). The highest
residual scores are on Indoor
($25.45$) and Kitchen ($18.97$), both compound-noun-heavy
categories where the compound-head guard (Sec.~\ref{sec:resolver})
is intentionally conservative. This pattern is consistent with
the ablation. Evasion gains come primarily from span grounding,
while dynamic binding mainly recovers preserve-side fidelity.

\begin{table}[!ht]
\centering
\tiny
\renewcommand{\arraystretch}{0.95}
\caption{\textbf{Per-category SEE Table~3 (evasion).} Lower is
better. Per-column best is \textbf{bold}, second-best
\underline{underlined}; \HGSS{} is the column-best on every
super-category and on the average.}
\label{tab:see_per_cat}
\setlength{\tabcolsep}{2pt}
\resizebox{\linewidth}{!}{%
\begin{tabular}{lcccccccccccc}
\toprule
Method & Vehicle & Outdoor & Animal & Access. & Sports & Kitchen & Food & Furn. & Elec. & Appl. & Indoor & Avg \\
\midrule
Unedited SD                          & 95.65 & 91.04 & 92.56 & 91.72 & 88.29 & 94.52 & 94.31 & 96.97 & 86.02 & 91.04 & 85.99 & 91.65 \\
UCE~\cite{gandikota2024uce}          & 94.19 & 89.54 & 94.81 & 81.60 & 83.43 & 63.12 & 89.83 & 97.02 & 81.05 & 90.55 & 61.56 & 84.25 \\
RECE~\cite{gong2024rece}             & 95.39 & 93.28 & 91.86 & 75.40 & 81.04 & 62.25 & 93.83 & 96.15 & 78.63 & 88.36 & 62.17 & 83.49 \\
MACE~\cite{lu2024mace}               & 91.55 & 88.93 & 89.68 & 77.88 & 82.02 & 58.87 & 88.01 & 93.64 & 78.91 & 87.83 & 57.38 & 81.34 \\
SPM~\cite{lyu2024spm}                & 94.82 & 91.18 & 92.71 & 79.13 & 84.51 & 65.70 & 87.93 & 89.43 & 79.92 & 90.97 & 59.92 & 83.29 \\
ESD~\cite{gandikota2023esd}          & 94.00 & 91.50 & 94.20 & 81.30 & 84.80 & 66.10 & 90.10 & 90.60 & 81.20 & 90.90 & 62.20 & 84.26 \\
AdvUnlearn~\cite{zhang2024advunlearn} & 93.20 & 90.90 & 93.10 & 80.30 & 84.00 & 64.90 & 89.70 & 90.10 & 80.40 & 90.60 & 60.70 & 83.45 \\
\SS{}~\cite{xiong2025ss}
                & \underline{21.09} & \underline{25.31} & \underline{33.75}
                & \underline{36.56} & \underline{30.47} & \underline{29.02}
                & \underline{25.31} & \underline{29.95} & \underline{35.16}
                & \underline{26.88} & \underline{31.47} & \underline{29.54} \\
\textbf{\HGSS{} (ours)}
                & \textbf{6.25}  & \textbf{16.25} & \textbf{1.56}
                & \textbf{15.63} & \textbf{4.37}  & \textbf{18.97}
                & \textbf{4.37}  & \textbf{4.69}  & \textbf{2.34}
                & \textbf{10.31} & \textbf{25.45} & \textbf{10.02} \\
\bottomrule
\end{tabular}}
\end{table}

\begin{figure}[!ht]
\centering
\includegraphics[width=0.74\textwidth]{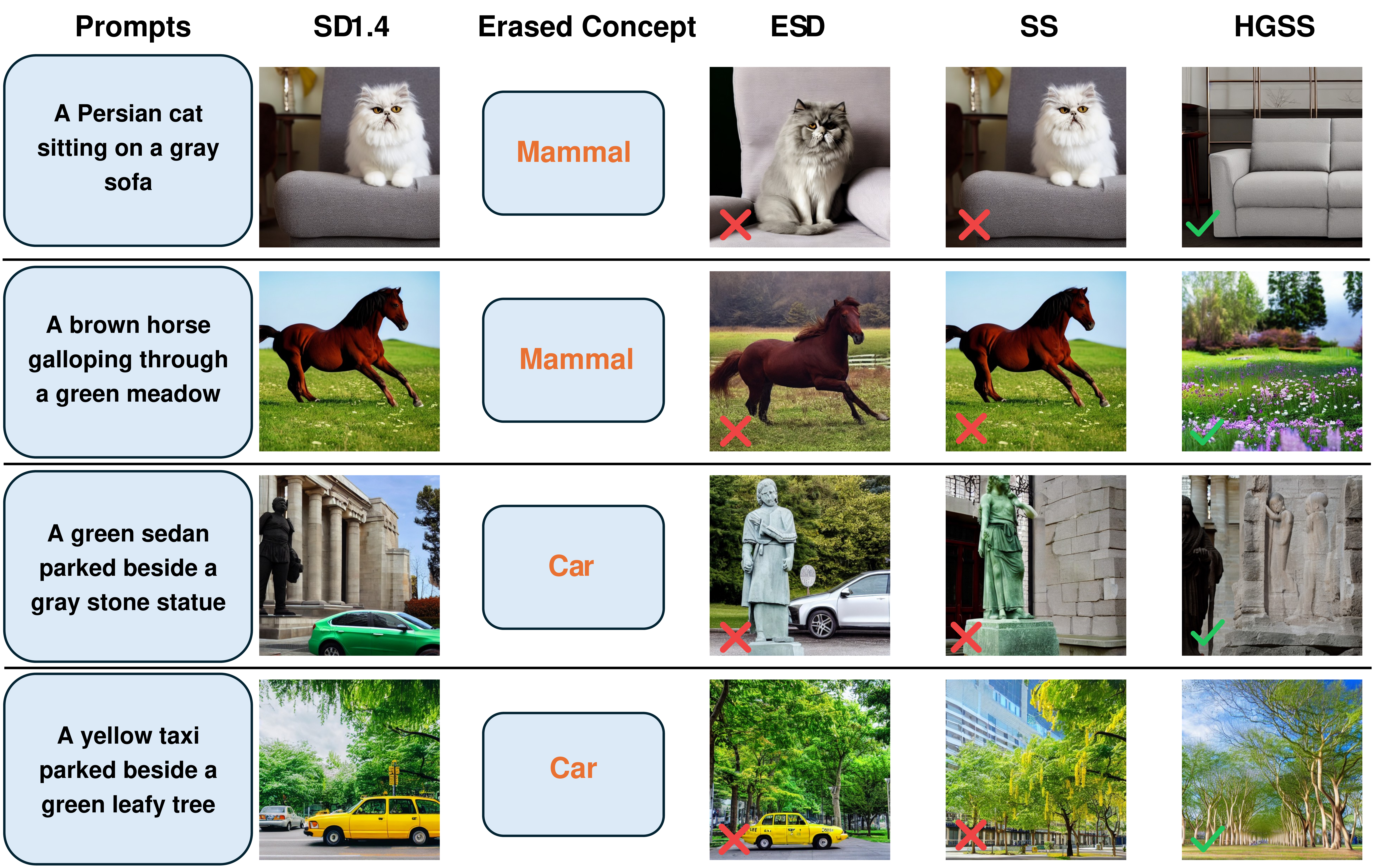}
\caption{\textbf{Qualitative analysis on four SEE prompts.}
Rows~1--2 show hierarchical evasion under erase target
``mammal'': ESD and \SS{} retain mammal descendants, while
\HGSS{} suppresses them. Rows~3--4 show car erasure with colour
attributes under erase target ``car''. \SS{} can retain the car
or transfer colour to preserved objects, while \HGSS{} removes
the car and preserves the non-target scene.}
\label{fig:qualitative_extended}
\end{figure}

\paragraph{Qualitative analysis.}
Figure~\ref{fig:qualitative_extended} provides a broader
qualitative analysis on four SEE prompts, complementing the
teaser in Figure~\ref{fig:teaser}. Rows~1--2 test hierarchical
evasion under target ``mammal'', and rows~3--4 test object
erasure with colour attributes under target ``car''.

\subsection{Main Results on UnlearnCanvas}
\label{sec:exp-uc-main}

Table~\ref{tab:uc_main} reports UnlearnCanvas in the standard
UA/IRA/CRA layout, with the six-metric effectiveness average,
COCO-30K FID, and resource columns. The evaluation covers all
$50$ official style targets and all $20$ object targets.
\HGSS{} reaches a six-metric average of $95.80$, exceeding the
matched \SS{} baseline by $0.31$. UC is used here to check generality. The average improves, but the per-metric
breakdown below shows a retain-side trade-off on object IRA/CRA.

\begin{table}[!ht]
\centering
\small
\caption{\textbf{UnlearnCanvas comparison.}
$\uparrow$: UA/IRA/CRA, Avg.; $\downarrow$: FID, Mem.~(GB),
Storage~(GB). \SS{} and \HGSS{} use matched runs on the full UC
set ($50$ style $+$ $20$ object targets). Best \textbf{bold} per
column; baseline FIDs are source-reported under a different
reference set, so no FID best is marked.}
\label{tab:uc_main}
\setlength{\tabcolsep}{2.6pt}
\resizebox{\linewidth}{!}{%
\renewcommand{\arraystretch}{0.95}%
\begin{tabular}{lcccccccccc}
\toprule
 & \multicolumn{3}{c}{Style Unlearning} & \multicolumn{3}{c}{Object Unlearning}
 & Effect. & \multicolumn{3}{c}{Efficiency}\\
\cmidrule(lr){2-4}\cmidrule(lr){5-7}\cmidrule(lr){8-8}\cmidrule(lr){9-11}
Method & UA & IRA & CRA & UA & IRA & CRA & Avg.
       & FID & Mem. & Stor.\\
\midrule
ESD~\cite{gandikota2023esd}        & \textbf{98.58} & 80.97 & 93.96 & 92.15 & 55.78 & 44.23 & 77.61 & 65.55 & 17.8 & 4.3 \\
FMN~\cite{zhang2024fmn}            & 88.48 & 56.77 & 46.60 & 45.64 & 90.63 & 73.46 & 66.93 & 131.37 & 17.9 & 4.2 \\
UCE~\cite{gandikota2024uce}        & 98.40 & 60.22 & 47.71 & \textbf{94.31} & 39.35 & 34.67 & 62.45 & 182.01 & 5.1 & 1.7 \\
CA~\cite{kumari2023ca}             & 60.82 & 96.01 & 92.70 & 46.67 & 90.11 & 81.97 & 78.05 & 54.21 & 10.1 & 4.2 \\
SalUn~\cite{fan2024salun}          & 86.26 & 90.39 & 95.08 & 86.91 & 96.35 & \textbf{99.59} & 92.43 & 61.05 & 30.8 & 4.0 \\
SEOT~\cite{li2024seot}             & 56.90 & 94.68 & 84.31 & 23.25 & 95.57 & 82.71 & 72.91 & 62.38 & 7.34 & \textbf{0.0} \\
SPM~\cite{lyu2024spm}              & 60.94 & 92.39 & 84.33 & 71.25 & 90.79 & 81.65 & 80.23 & 59.79 & 6.9 & \textbf{0.0} \\
SHS~\cite{wu2024shs}               & 95.84 & 80.42 & 43.27 & 80.73 & 81.15 & 67.99 & 74.90 & 119.34 & 31.2 & 4.0 \\
EraseDiff~\cite{wu2024erasediff}   & 92.42 & 73.91 & 98.93 & 86.67 & 94.03 & 48.48 & 82.41 & 81.42 & 27.8 & 4.0 \\
SAeUron~\cite{cywinski2025saeuron} & 95.80 & \textbf{99.10} & \textbf{99.40} & 78.82 & 95.47 & 95.58 & 94.03 & 62.15 & \textbf{2.8} & 0.2 \\
\midrule
\SS{}~\cite{xiong2025ss}    & 92.74 & 94.93 & 97.72 & 90.16 & \textbf{98.27} & 99.15 & 95.49 & 56.77 & 7.2 & \textbf{0.0} \\
\textbf{\HGSS{}} (ours)            & 95.28 & 97.28 & 97.78 & 94.27 & 94.63 & 95.58 & \textbf{95.80} & 56.15 & 7.2 & \textbf{0.0} \\
\bottomrule
\end{tabular}}
\end{table}

The gains concentrate on unlearning accuracy rather than uniformly
improving all retention proxies. Style UA improves from $92.74$ to
$95.28$ ($+2.54$), style IRA from $94.93$ to $97.28$ ($+2.35$),
and object UA from $90.16$ to $94.27$ ($+4.11$). Two of the six
metrics, object IRA ($98.27 \to 94.63$) and object CRA
($99.15 \to 95.58$), decrease by $3$--$4$pp. This suggests that
the SEE-oriented binding objective does not uniformly improve
every classifier-based retention proxy in UC. Gains in style
UA/IRA/CRA and object UA together offset the two object-retention
losses. So \HGSS{} is ahead of the matched \SS{} baseline by
$0.31$ on the six-metric average. The COCO-30K FID of $56.15$
improves over the matched \SS{} baseline ($56.77$) by $0.62$
points.

\subsection{Direct Lexical Safety Erasure on I2P}
\label{sec:exp-i2p}

I2P evaluates direct lexical unsafe-content removal, a different
operating regime from the compositional setting studied here, and
is therefore reported separately. \HGSS{} keeps the same frozen
checkpoint and unified inference-time framework as on SEE and
UnlearnCanvas. For direct safety targets, the coarse edit anchors
on the requested safety concept; benign prompt content is not
added to the erase set.

\begin{table}[!ht]
\centering
\footnotesize
\renewcommand{\arraystretch}{0.95}
\caption{\textbf{Direct lexical safety erasure on I2P
(representative methods).} Total $=$ NudeNet detections over eight
body-part classes on $4{,}703$ I2P prompts (\texttt{paper8 +
threshold=0.6}; lower is safer). FID, CLIP are MS-COCO-30K.
SD~v1.4 is no-intervention; \textbf{bold} excludes SD. Full
$12$-method breakdown in App.~C.3,
Suppl.\ Table~2.}
\label{tab:i2p_main}
\setlength{\tabcolsep}{5pt}
\begin{tabular}{llccc}
\toprule
Method & Regime & Total$\downarrow$ & FID$\downarrow$ & CLIP$\uparrow$ \\
\midrule
SD v1.4~\cite{rombach2022ldm}     & Base          & 751         & 14.04          & 31.34          \\
ESD-u~\cite{gandikota2023esd}     & Weight-edit   & 55          & 15.10          & 30.21          \\
UCE~\cite{gandikota2024uce}       & Weight-edit   & 165         & 14.07          & 30.85          \\
MACE~\cite{lu2024mace}            & Weight-edit   & 123         & \textbf{13.42} & 29.41          \\
Receler~\cite{huang2024receler}   & Weight-edit   & 159         & 14.10          & \textbf{31.02} \\
SAFREE~\cite{yoon2025safree}      & Training-free & 82          & --             & --             \\
\SS{}~\cite{xiong2025ss}          & Training-free & 46          & 13.45          & 30.71          \\
\textbf{\HGSS{}} (ours)           & Training-free & \textbf{25} & 14.56\textsuperscript{\dag} & 30.19\textsuperscript{\dag} \\
\bottomrule
\end{tabular}
\\[0.3em]
\parbox{0.96\linewidth}{\footnotesize
\textsuperscript{\dag}\HGSS{} FID/CLIP under default config;
target-specific anchoring inactive on MS-COCO prompts.}
\end{table}

Table~\ref{tab:i2p_main} reports Total $=25$ unsafe detections for
\HGSS{}, lower than all seven representative I2P baselines
shown and the lowest in the full $12$-method comparison
(App.~C.3, Suppl.\ Table~2). This is a direct lexical safety
check rather than evidence for compositional erasure. \HGSS{}
retains safety-erasure behavior at the standard I2P operating
point.
COCO-30K FID and CLIP fall within the range spanned by the other
baselines (Suppl.\ Table~2).

\subsection{Module Ablation}
\label{sec:exp-ablate}

Each \HGSS{} module's contribution was isolated through the
$2{\times}2$ factorial ablation in Table~\ref{tab:ablation_2x2},
evaluated on the full SEE benchmark at $\beta{=}{-}0.12$ across
all $11$ super-categories. The ablation used the same coarse
text-side edit interface in every row and toggled only the two
modules introduced in Sec.~\ref{sec:resolver} and
Sec.~\ref{sec:dyntext}.

\begin{table}[!ht]
\centering
\small
\caption{\textbf{Full SEE module ablation} at $\beta{=}{-}0.12$.
Lower is better for Neighbor $E$, Evasion, Attr$E$, and Attr$P$;
higher is better for Neighbor $P$. Best per column in
\textbf{bold}.}
\label{tab:ablation_2x2}
\setlength{\tabcolsep}{3pt}
\renewcommand{\arraystretch}{0.85}
\resizebox{\linewidth}{!}{%
\begin{tabular}{lccccc}
\toprule
Variant &
Neighbor $E$\,$\downarrow$ &
Neighbor $P$\,$\uparrow$ &
Evasion\,$\downarrow$ &
Attr$E$\,$\downarrow$ &
Attr$P$\,$\downarrow$ \\
\midrule
Base Coarse Surgery
  & 25.30 & 31.22 & 29.54 & 80.41 & 82.22 \\
Base + Hierarchical Span Grounding
  & 9.88 & 21.39 & 12.19 & \textbf{35.21} & 42.44 \\
Base + Dynamic Attribute Binding
  & 24.71 & \textbf{31.31} & 29.40 & 71.96 & 72.01 \\
\textbf{\HGSS{} full}
  & \textbf{8.08} & 30.04 & \textbf{10.02} & 41.44 & \textbf{38.30} \\
\bottomrule
\end{tabular}}
\end{table}

\paragraph{Module attribution.}
Table~\ref{tab:ablation_2x2} shows a clean division of labour.
\textbf{Grounding improves erasure}, cutting Neighbor~$E$, Evasion,
and Attr$P$ substantially over Base Coarse Surgery at the cost of
preserve-side Neighbor~$P$. \textbf{Binding alone is not erasure}.
Without improved routing it cannot correct missed erase spans.
Stacked on the resolver, it recovers Neighbor~$P$ and Attr$P$
while keeping the erasure gains, trading a small amount of
resolver-only Attr$E$ for the intended preserve-side recovery.

\section{Conclusion}
\label{sec:conclusion}

This paper introduced \HGSS{}, a training-free correction to
text-side concept erasure for text-to-image diffusion models.
Compositional erasure needs both target routing and preserve-side
binding control. Hierarchical span grounding (what to edit) and
dynamic attribute binding (how the edit evolves) together cast
erasure as the inverse objective to compositional generation.
Experiments on SEE and UnlearnCanvas support this view.

\bibliography{egbib}

\begin{thebibliography}{41}
\providecommand{\natexlab}[1]{#1}
\providecommand{\url}[1]{\texttt{#1}}
\expandafter\ifx\csname urlstyle\endcsname\relax
  \providecommand{\doi}[1]{doi: #1}\else
  \providecommand{\doi}{doi: \begingroup \urlstyle{rm}\Url}\fi

\bibitem[Bui et~al.(2025)Bui, Vu, Vuong, Le, Montague, Abraham, Kim, and Phung]{bui2025age}
Anh~Tuan Bui, Thuy-Trang Vu, Long~Tung Vuong, Trung Le, Paul Montague, Tamas Abraham, Junae Kim, and Dinh Phung.
\newblock Fantastic targets for concept erasure in diffusion models and where to find them.
\newblock In \emph{ICLR}, 2025.

\bibitem[Chavhan et~al.(2025)Chavhan, Li, and Hospedales]{chavhan2025conceptprune}
Ruchika Chavhan, Da~Li, and Timothy~M. Hospedales.
\newblock {ConceptPrune}: Concept editing in diffusion models via skilled neuron pruning.
\newblock In \emph{ICLR}, 2025.

\bibitem[Chefer et~al.(2023)Chefer, Alaluf, Vinker, Wolf, and Cohen-Or]{chefer2023attendexcite}
Hila Chefer, Yuval Alaluf, Yael Vinker, Lior Wolf, and Daniel Cohen-Or.
\newblock {Attend-and-Excite}: Attention-based semantic guidance for text-to-image diffusion models.
\newblock \emph{ACM Transactions on Graphics (SIGGRAPH)}, 42\penalty0 (4), 2023.
\newblock arXiv:2301.13826.

\bibitem[Chin et~al.(2024)Chin, Jiang, Huang, Chen, and Chiu]{chin2024p4d}
Zhi-Yi Chin, Chieh-Ming Jiang, Ching-Chun Huang, Pin-Yu Chen, and Wei-Chen Chiu.
\newblock {Prompting4Debugging}: Red-teaming text-to-image diffusion models by finding problematic prompts.
\newblock In \emph{ICML}, 2024.

\bibitem[Cywi{\'n}ski and Deja(2025)]{cywinski2025saeuron}
Bartosz Cywi{\'n}ski and Kamil Deja.
\newblock {SAeUron}: Interpretable concept unlearning in diffusion models with sparse autoencoders.
\newblock In \emph{ICML}, 2025.
\newblock arXiv:2501.18052.

\bibitem[Fan et~al.(2024)Fan, Liu, Zhang, Wong, Wei, and Liu]{fan2024salun}
Chongyu Fan, Jiancheng Liu, Yihua Zhang, Eric Wong, Dennis Wei, and Sijia Liu.
\newblock {SalUn}: Empowering machine unlearning via gradient-based weight saliency in both image classification and generation.
\newblock In \emph{ICLR}, 2024.

\bibitem[Feng et~al.(2023)Feng, He, Fu, Jampani, Akula, Narayana, Basu, Wang, and Wang]{feng2023structurediffusion}
Weixi Feng, Xuehai He, Tsu-Jui Fu, Varun Jampani, Arjun~R. Akula, Pradyumna Narayana, Sugato Basu, Xin~Eric Wang, and William~Yang Wang.
\newblock Training-free structured diffusion guidance for compositional text-to-image synthesis.
\newblock In \emph{ICLR}, 2023.

\bibitem[Gandikota et~al.(2023)Gandikota, Materzy{\'n}ska, Fiotto-Kaufman, and Bau]{gandikota2023esd}
Rohit Gandikota, Joanna Materzy{\'n}ska, Jaden Fiotto-Kaufman, and David Bau.
\newblock Erasing concepts from diffusion models.
\newblock In \emph{ICCV}, 2023.

\bibitem[Gandikota et~al.(2024)Gandikota, Orgad, Belinkov, Materzy{\'n}ska, and Bau]{gandikota2024uce}
Rohit Gandikota, Hadas Orgad, Yonatan Belinkov, Joanna Materzy{\'n}ska, and David Bau.
\newblock Unified concept editing in diffusion models.
\newblock In \emph{WACV}, 2024.

\bibitem[Gong et~al.(2024)Gong, Chen, Wei, Chen, and Jiang]{gong2024rece}
Chao Gong, Kai Chen, Zhipeng Wei, Jingjing Chen, and Yu-Gang Jiang.
\newblock Reliable and efficient concept erasure of text-to-image diffusion models.
\newblock In \emph{ECCV}, 2024.
\newblock arXiv:2407.12383.

\bibitem[Hertz et~al.(2023)Hertz, Mokady, Tenenbaum, Aberman, Pritch, and Cohen-Or]{hertz2023p2p}
Amir Hertz, Ron Mokady, Jay Tenenbaum, Kfir Aberman, Yael Pritch, and Daniel Cohen-Or.
\newblock Prompt-to-prompt image editing with cross-attention control.
\newblock In \emph{ICLR}, 2023.

\bibitem[Ho and Salimans(2022)]{ho2022cfg}
Jonathan Ho and Tim Salimans.
\newblock Classifier-free diffusion guidance.
\newblock \emph{arXiv preprint arXiv:2207.12598}, 2022.
\newblock Short version: NeurIPS 2021 Workshop on Deep Generative Models and Downstream Applications.

\bibitem[Ho et~al.(2020)Ho, Jain, and Abbeel]{ho2020ddpm}
Jonathan Ho, Ajay Jain, and Pieter Abbeel.
\newblock Denoising diffusion probabilistic models.
\newblock In \emph{NeurIPS}, 2020.
\newblock arXiv:2006.11239.

\bibitem[Huang et~al.(2024)Huang, Chang, Tsai, Lai, Yang, and Wang]{huang2024receler}
Chi-Pin Huang, Kai-Po Chang, Chung-Ting Tsai, Yung-Hsuan Lai, Fu-En Yang, and Yu-Chiang~Frank Wang.
\newblock {Receler}: Reliable concept erasing of text-to-image diffusion models via lightweight erasers.
\newblock In \emph{ECCV}, 2024.
\newblock arXiv:2311.17717.

\bibitem[Jain et~al.(2024)Jain, Kobayashi, Shibuya, Takida, Memon, Togelius, and Mitsufuji]{jain2024trasce}
Anubhav Jain, Yuya Kobayashi, Takashi Shibuya, Yuhta Takida, Nasir~D. Memon, Julian Togelius, and Yuki Mitsufuji.
\newblock {TraSCE}: Trajectory steering for concept erasure.
\newblock \emph{arXiv preprint arXiv:2412.07658}, 2024.

\bibitem[Kumari et~al.(2023)Kumari, Zhang, Wang, Shechtman, Zhang, and Zhu]{kumari2023ca}
Nupur Kumari, Bingliang Zhang, Sheng-Yu Wang, Eli Shechtman, Richard Zhang, and Jun-Yan Zhu.
\newblock Ablating concepts in text-to-image diffusion models.
\newblock In \emph{ICCV}, 2023.

\bibitem[Lee et~al.(2025{\natexlab{a}})Lee, Lim, and Chun]{lee2025gloce}
Byung~Hyun Lee, Sungjin Lim, and Se~Young Chun.
\newblock Localized concept erasure for text-to-image diffusion models using training-free gated low-rank adaptation.
\newblock In \emph{CVPR}, 2025{\natexlab{a}}.
\newblock arXiv:2503.12356.

\bibitem[Lee et~al.(2025{\natexlab{b}})Lee, Lim, Lee, Kang, and Chun]{lee2025cpe}
Byung~Hyun Lee, Sungjin Lim, Seunggyu Lee, Dong~Un Kang, and Se~Young Chun.
\newblock Concept pinpoint eraser for text-to-image diffusion models via residual attention gate.
\newblock In \emph{ICLR}, 2025{\natexlab{b}}.

\bibitem[Li et~al.(2024)Li, van~de Weijer, Hu, Khan, Hou, Wang, and Yang]{li2024seot}
Senmao Li, Joost van~de Weijer, Taihang Hu, Fahad~Shahbaz Khan, Qibin Hou, Yaxing Wang, and Jian Yang.
\newblock Get what you want, not what you don't: Image content suppression for text-to-image diffusion models.
\newblock In \emph{ICLR}, 2024.
\newblock arXiv:2402.05375.

\bibitem[Lu et~al.(2024)Lu, Wang, Li, Liu, and Kong]{lu2024mace}
Shilin Lu, Zilan Wang, Leyang Li, Yanzhu Liu, and Adams Wai-Kin Kong.
\newblock {MACE}: Mass concept erasure in diffusion models.
\newblock In \emph{CVPR}, 2024.

\bibitem[Lyu et~al.(2024)Lyu, Yang, Hong, Chen, Jin, He, Xue, Han, and Ding]{lyu2024spm}
Mengyao Lyu, Yuhong Yang, Haiwen Hong, Hui Chen, Xuan Jin, Yuan He, Hui Xue, Jungong Han, and Guiguang Ding.
\newblock One-dimensional adapter to rule them all: Concepts, diffusion models, and erasing applications.
\newblock In \emph{CVPR}, pages 7559--7568, 2024.
\newblock arXiv:2312.16145.

\bibitem[Miller(1995)]{miller1995wordnet}
George~A. Miller.
\newblock {WordNet}: A lexical database for english.
\newblock \emph{Communications of the ACM}, 38\penalty0 (11):\penalty0 39--41, 1995.

\bibitem[Na et~al.(2025)Na, Park, Sim, Shin, Bae, Kang, Kwon, Kang, and Moon]{na2025date}
Byeonghu Na, Minsang Park, Gyuwon Sim, Donghyeok Shin, HeeSun Bae, Mina Kang, Se~Jung Kwon, Wanmo Kang, and Il-Chul Moon.
\newblock {Diffusion Adaptive Text Embedding} for text-to-image diffusion models.
\newblock In \emph{NeurIPS}, 2025.
\newblock arXiv:2510.23974.

\bibitem[Nie et~al.(2025)Nie, Yao, Liu, Wang, and Bian]{nie2025cogfd}
Hongyi Nie, Quanming Yao, Yang Liu, Zhen Wang, and Yatao Bian.
\newblock Erasing concept combination from text-to-image diffusion model.
\newblock In \emph{ICLR}, 2025.

\bibitem[Radford et~al.(2021)Radford, Kim, Hallacy, Ramesh, Goh, Agarwal, Sastry, Askell, Mishkin, Clark, Krueger, and Sutskever]{radford2021clip}
Alec Radford, Jong~Wook Kim, Chris Hallacy, Aditya Ramesh, Gabriel Goh, Sandhini Agarwal, Girish Sastry, Amanda Askell, Pamela Mishkin, Jack Clark, Gretchen Krueger, and Ilya Sutskever.
\newblock Learning transferable visual models from natural language supervision.
\newblock In \emph{ICML}, 2021.

\bibitem[Rassin et~al.(2023)Rassin, Hirsch, Glickman, Ravfogel, Goldberg, and Chechik]{rassin2023syngen}
Royi Rassin, Eran Hirsch, Daniel Glickman, Shauli Ravfogel, Yoav Goldberg, and Gal Chechik.
\newblock Linguistic binding in diffusion models: Enhancing attribute correspondence through attention map alignment.
\newblock In \emph{NeurIPS}, 2023.

\bibitem[Rombach et~al.(2022)Rombach, Blattmann, Lorenz, Esser, and Ommer]{rombach2022ldm}
Robin Rombach, Andreas Blattmann, Dominik Lorenz, Patrick Esser, and Bj{\"o}rn Ommer.
\newblock High-resolution image synthesis with latent diffusion models.
\newblock In \emph{CVPR}, 2022.

\bibitem[Saha et~al.(2025)Saha, Saha, Gaur, and Gokhale]{see2025}
Shaswati Saha, Sourajit Saha, Manas Gaur, and Tejas Gokhale.
\newblock Side effects of erasing concepts from diffusion models.
\newblock In \emph{Findings of the Association for Computational Linguistics: EMNLP 2025}, pages 14991--15007, 2025.
\newblock arXiv:2508.15124.

\bibitem[Schramowski et~al.(2023)Schramowski, Brack, Deiseroth, and Kersting]{schramowski2023sld}
Patrick Schramowski, Manuel Brack, Bj{\"o}rn Deiseroth, and Kristian Kersting.
\newblock Safe latent diffusion: Mitigating inappropriate degeneration in diffusion models.
\newblock In \emph{CVPR}, 2023.

\bibitem[Song et~al.(2021)Song, Meng, and Ermon]{song2021ddim}
Jiaming Song, Chenlin Meng, and Stefano Ermon.
\newblock Denoising diffusion implicit models.
\newblock In \emph{ICLR}, 2021.

\bibitem[Tsai et~al.(2024)Tsai, Hsu, Xie, Lin, Chen, Li, Chen, Yu, and Huang]{tsai2024ringabell}
Yu-Lin Tsai, Chia-Yi Hsu, Chulin Xie, Chih-Hsun Lin, Jia~You Chen, Bo~Li, Pin-Yu Chen, Chia-Mu Yu, and Chun-Ying Huang.
\newblock {Ring-A-Bell!} how reliable are concept removal methods for diffusion models?
\newblock In \emph{ICLR}, 2024.
\newblock arXiv:2310.10012.

\bibitem[Wang et~al.(2025)Wang, Li, Mu, Hao, Liu, Wang, and He]{wang2025valuespace}
Yuan Wang, Ouxiang Li, Tingting Mu, Yanbin Hao, Kuien Liu, Xiang Wang, and Xiangnan He.
\newblock Precise, fast, and low-cost concept erasure in value space: Orthogonal complement matters.
\newblock In \emph{CVPR}, 2025.

\bibitem[Wu and Harandi(2024)]{wu2024shs}
Jing Wu and Mehrtash Harandi.
\newblock Scissorhands: Scrub data influence via connection sensitivity in networks.
\newblock In \emph{ECCV}, 2024.
\newblock arXiv:2401.06187.

\bibitem[Wu et~al.(2025)Wu, Le, Hayat, and Harandi]{wu2024erasediff}
Jing Wu, Trung Le, Munawar Hayat, and Mehrtash Harandi.
\newblock Erasing undesirable influence in diffusion models.
\newblock In \emph{CVPR}, pages 28263--28273, 2025.
\newblock arXiv:2401.05779.

\bibitem[Xiong et~al.(2025)Xiong, Liu, Ye, Liu, and Xu]{xiong2025ss}
Lexiang Xiong, Chengyu Liu, Jingwen Ye, Yan Liu, and Yuecong Xu.
\newblock Semantic surgery: Zero-shot concept erasure in diffusion models.
\newblock In \emph{NeurIPS}, 2025.
\newblock arXiv:2510.22851.

\bibitem[Yang et~al.(2024)Yang, Gao, Wang, Ho, Xu, and Xu]{yang2024mmadiffusion}
Yijun Yang, Ruiyuan Gao, Xiaosen Wang, Tsung-Yi Ho, Nan Xu, and Qiang Xu.
\newblock {MMA-Diffusion}: Multimodal attack on diffusion models.
\newblock In \emph{CVPR}, 2024.
\newblock arXiv:2311.17516.

\bibitem[Yoon et~al.(2025)Yoon, Yu, Patil, Yao, and Bansal]{yoon2025safree}
Jaehong Yoon, Shoubin Yu, Vaidehi Patil, Huaxiu Yao, and Mohit Bansal.
\newblock {SAFREE}: Training-free and adaptive guard for safe text-to-image and video generation.
\newblock In \emph{ICLR}, 2025.
\newblock arXiv:2410.12761.

\bibitem[Zhang et~al.(2024{\natexlab{a}})Zhang, Wang, Xu, Wang, and Shi]{zhang2024fmn}
Gong Zhang, Kai Wang, Xingqian Xu, Zhangyang Wang, and Humphrey Shi.
\newblock Forget-me-not: Learning to forget in text-to-image diffusion models.
\newblock In \emph{CVPR Workshops}, pages 1755--1764, 2024{\natexlab{a}}.

\bibitem[Zhang et~al.(2024{\natexlab{b}})Zhang, Fan, Zhang, Yao, Jia, Liu, Zhang, Liu, Kompella, Liu, and Liu]{zhang2024unlearncanvas}
Yihua Zhang, Chongyu Fan, Yimeng Zhang, Yuguang Yao, Jinghan Jia, Jiancheng Liu, Gaoyuan Zhang, Gaowen Liu, Ramana~Rao Kompella, Xiaoming Liu, and Sijia Liu.
\newblock {UnlearnCanvas}: Stylized image dataset for enhanced machine unlearning evaluation in diffusion models.
\newblock In \emph{NeurIPS Datasets and Benchmarks Track}, 2024{\natexlab{b}}.

\bibitem[Zhang et~al.(2024{\natexlab{c}})Zhang, Chen, Jia, Zhang, Fan, Liu, Hong, Ding, and Liu]{zhang2024advunlearn}
Yimeng Zhang, Xin Chen, Jinghan Jia, Yihua Zhang, Chongyu Fan, Jiancheng Liu, Mingyi Hong, Ke~Ding, and Sijia Liu.
\newblock Defensive unlearning with adversarial training for robust concept erasure in diffusion models.
\newblock In \emph{NeurIPS}, 2024{\natexlab{c}}.
\newblock arXiv:2405.15234.

\bibitem[Zhang et~al.(2024{\natexlab{d}})Zhang, Jia, Chen, Chen, Zhang, Liu, Ding, and Liu]{zhang2024unlearndiffatk}
Yimeng Zhang, Jinghan Jia, Xin Chen, Aochuan Chen, Yihua Zhang, Jiancheng Liu, Ke~Ding, and Sijia Liu.
\newblock To generate or not? safety-driven unlearned diffusion models are still easy to generate unsafe images \ldots for now.
\newblock In \emph{ECCV}, 2024{\natexlab{d}}.

\end{thebibliography}

\end{document}